# テキスト・画像による情報消化効率のメディア別構成因子抽出

小池 央晟[1] 早矢仕 晃章[2]
[1,2] 東京大学 工学部 システム創成学科
[1] koike-hiroaki-uts@g.ecc.u-tokyo.ac.jp
[2] hayashi@sys.t.u-tokyo.ac.jp

## 概要

情報通信技術の発展と普及は一度に発信可能な情報量の増大と多様化をもたらした．しかし，情報量の増大や情報の取捨選択が必ずしも情報の理解を促進する訳ではない．また，従来の情報伝達評価では受信者への到達までが対象であり，本来の目的である受信者の情報取得後の理解は十分に考慮されてこなかった．本研究では受信者が取得した情報，その内容及び趣旨を正しく理解することを指す"情報の消化"概念を提案する．実験では，階層的因子分析を用いて情報の消化効率の評価モデルを提案し，メディア別に消化しやすさを構成する因子を抽出した．

## 1 はじめに

情報伝達の工学的評価には，通信技術の比較を目的として情報伝播モデル等が用いられてきた．近年，個々の情報の伝達力評価も試みられている[1]が，その主目的は情報が取得されやすくなる度合いの定量評価であり，情報の到達を目的とする課題の解決には力点が置かれていない．情報量の増加に伴う課題[2]に対応するには受信者による情報の内容/趣旨の正しい理解を含めた定量評価方法が求められる．

一方，情報取得後の理解は教育心理学分野[3]で研究され，既存知識と新規情報理解の間には強い相関関係があることが分かっている．しかし，事前知識が形成する概念フレームワークの変化の正答率等を計測する手法が主であり，情報自体の性質やメディアの差異による情報理解の分析は十分ではない．実社会における情報伝達の議論には，情報自体の性質から情報の受信者・発信者各位との関係性を踏まえた分析と定量的な評価モデルが喫緊の課題である．

本研究のゴールは，情報発信者の趣旨・意図から受信者がそれを理解するまでを含めた情報伝達のプロセス全体の情報伝達効率の定量評価手法の構築である．この定量評価にあたり，まず本論文では受信者が取得した情報，その内容及び趣旨を正しく理解することを指す"情報の消化"概念を提案する．実験では，階層的因子分析を用いて情報の消化効率の評価モデルを提案し，メディア別に消化しやすさを構成する因子を抽出した．

## 2 提案手法

### 2.1 情報の消化

本研究ではファジィな概念である情報の中でも「発信後・受信者が理解する過程で変更されず，1回の発信で完結する」ものを研究対象とした．すなわち，複数回の質疑応答を重ねて理解を深めるような情報伝達や，受信者が情報を閲覧中に五月雨式に情報が追加されていくような情報は対象に含まない．つまり，連続するチャット/スレッド形式での情報発信やテレビのニュース番組等の動的な情報発信は含まない．これらは時間連続的・群的な境界が定まらず，定量評価が困難であるためである．

情報の消化とは，受信者が取得した情報及び情報の内容・趣旨を正しく理解することを指す．ある情報の受信者への到達と，受信者の実際の理解とは区別される．

### 2.2 情報の消化効率定量評価モデル

情報の消化の概念の導入により，ある情報を相手に伝え理解してもらうという情報伝達の本来の目的を踏まえた情報伝達効率の評価が可能となる．情報の消化効率 ($\rho$) を式(1)で表す．$\omega_{\text{group}}(i)$は$i$番目のグループ因子の信頼度，$F_i$がデータの変動を説明できる割合のうち一般因子の影響を排した割合を指す．そして，各グループ因子の評価を$\text{ev}(F_i)$と表す．

$$\rho = \sum_i \frac{\omega_{\text{group}}(i) \cdot \text{ev}(F_i)}{\sum_{j=0}^{n} \omega_{\text{group}}(j)} \quad (1)$$

これに因子分析で抽出されたグループ因子を適用すれば，メディア別の定量評価式となる．すなわち，3.2節で後述する4件のメディア A: オンラインニュース記事, B: オンライン広告, C: ネットショッピン

グ（ECサイトの商品ページ），D: 論文/レポートに対し，以下の式(2)が得られる．

$$\rho(A) = \text{ev}(未開拓性) \cdot 0.272 + \text{ev}(簡潔性) \cdot 0.360 + \text{ev}(親密性) \cdot 0.378$$
$$\rho(B) = \text{ev}(網羅性) \cdot 0.384 + \text{ev}(簡潔性) \cdot 0.312 + \text{ev}(結論へのアクセス性) \cdot 0.304 \quad (2)$$
$$\rho(C) = \text{ev}(親密性) \cdot 0.279 + \text{ev}(簡潔性) \cdot 0.400 + \text{ev}(未開拓性) \cdot 0.321$$
$$\rho(D) = \text{ev}(説明可能性) \cdot 0.557 + \text{ev}(簡潔性) \cdot 0.443$$

なお式中の各グループ因子（$F_i$）の評価については，各因子を支える観測変数を参考に情報がどれ程その性質を備えているかを0~1の値で複数名による主観評価を行い，平均値を採用する．

# 3 実験

## 3.1 目的

本研究の目的は，情報の消化効率の定量評価モデルの提案と評価である．そのため，実験では4件のメディアを対象に22件の観測変数に対して6段階の評価データを取得した．このデータを用いて階層的因子分析を行い，情報の消化に影響する要素，要素間の関係の計測・モデル化及び検証を行った．

## 3.2 データ取得

情報伝達を情報自体の性質，情報と発信者の関係，情報と受信者の関係の3要素を軸に，11件の特徴を選出した．そして，「メイントピックで説明・述べられている人/ものを既に知っている」，「メイントピックで説明・述べられている人/ものへの事前知識や先入観がない」といった反対の意味を成す組を一対とし，22件の観測変数を作った（表1）．これは，反対の意味を成す特徴は，必ずしも情報の消化に真逆の影響をもたらすとは限らないという仮説に基づく．「色やデザインによる装飾がある」という特徴に対し，「非常にそう思う」と評価されても，逆の意味である「色やデザインによる装飾がない」への評価は「どちらかといえば逆効果だと思う」と評価される可能性がある．実験では，各特徴について対となる質問を行うことで，情報の消化に真逆の影響をもたらすとは限らないということを確認する．

実験では，A: オンラインニュース記事, B: オンライン広告, C: ネットショッピング（ECサイトの商品ページ）, D: 論文/レポートの代表的な4件の情報伝達媒体（メディア）を対象とした．続いて，調査会社プラグのモニター16,000人に予備調査を行い，一定以上のメディア接触頻度の実験参加者を収集した．最終的にA, B, Cを400件，Dを100件取得するアンケート調査を実施した．アンケートでは回答者には「（メディア名）について，以下にあげたような**特徴があると**，その情報が**理解**しやすく，**後で誰かに伝えられるようになりやすい**ですか」と質問し，各特徴が情報の消化しやすさに与える影響について「非常にそう思う」〜「非常に逆効果だと思う」の片側3段階，計6段階で自己評価してもらった．

## 3.3 階層的因子分析（SEM分析）

分析ではメディア別に情報の消化に影響する因子を抽出した．分析手法には因子分析と重回帰分析を組み合わせたSEM分析を用いた．SEM分析は因子間の階層構造を検証できる．モデルには，全観測変数に影響を及ぼす一般因子（$g$）と，特定の観測変数のクラスターに影響を及ぼすグループ因子（$F_i$）の2層によって説明する双因子モデルを用いた[4]．一般因子は測定対象の広範で全体的な構成概念として，グループ因子はその下位領域の狭い構成概念として抽出されるため，階層性やグループ因子の独立性を検証できる．分析では因子数を仮定して与え，以下の評価指標の結果に基づき検討する[5]．

- 内的整合性: $\omega_{\text{total}}$
- 一次元性: ECV
- 適合度: fit
- $i$番目のグループ因子の信頼度: $\omega_{\text{group}}(i)$
- 一般因子の信頼度: $\omega_{\text{general}}$

$\omega_{\text{total}}$は全因子により変動の何割を説明できたかを示す．ECV (Explained Common Variance) は，共通因子で説明される分散を因子全体が説明できる割合（%）を示し，適合度はモデルのデータへの当てはまりの良さを示す．また$\omega_{\text{general}}$, $\omega_{\text{group}}(i)$は$\omega_{\text{total}}$の内，一般因子，$i$番目のグループ因子の信頼度によってそれぞれ説明される度合いを表す．

本実験ではこれらの指標を用いて最終的な因子数を決定し，モデルの階層性や，因子と各観測変数との関連の強さを示す因子負荷を参照しつつ各因子が表す概念を解釈する．なお本研究の特徴として，同一の観測変数群の利用により他メディア間の比較も可能なため，メディア間での比較も行いつつ因子を解釈した．

表 1　22 件の観測変数

| # | 観測変数 |
|---|---|
| 1 | 含まれるトピックや項目の数 が３つ以下 |
| 2 | 含まれるトピックや項目の数 が３つより多い |
| 3 | 多少情報量が多くなっても、もれなくダブりなく情報と根拠となるデータが提示されている |
| 4 | 情報や根拠となるデータがかなり省略されていても、意味合いや結論が端的にまとまっている |
| 5 | 情報の構成が多少複雑になっても、見出しや項目が細かく分かれている |
| 6 | 見出しや項目ごとにまとまっていなくても、情報の構成が単純である |
| 7 | 重要な部分が強く目立つよう、色やデザインによる装飾がされている |
| 8 | 重要な部分が多少目立たなくても、過剰な装飾がない |
| 9 | メイントピックで説明・述べられている内容（人/もの）を既に知っている |
| 10 | メイントピックで説明・述べられている内容（人/もの）の事前知識や先入観がない |
| 11 | メイントピック自体は知らなかったが、知っている要素(単語/人/場所など)の割合が高い |
| 12 | メイントピックも説明も、新しく知る要素(単語/人/場所など)の割合が高い |
| 13 | メイントピックが属するジャンル全体について、元々詳しく知っている |
| 14 | メイントピックが属するジャンル全体について、事前知識や先入観がない |
| 15 | テキスト・画像・動画など複数のタイプの情報が含まれている |
| 16 | テキスト・画像・動画のどれか１種類の情報に限定されている |
| 17 | 多少情報量が多くなっても、注釈・補足説明・関連情報などが多い |
| 18 | 注釈・補足説明・関連情報などが省かれ、端的にまとまっている |
| 19 | 多少情報量が多くなっても、情報発信者の立場や所属団体・企業に関する情報が含まれている |
| 20 | 情報の発信者の立場や所属団体・企業に関する情報が省かれ、端的にまとまっている |
| 21 | 情報発信者が発信する目的・メリット・意図が明記されている |
| 22 | 情報発信者が発信する目的・メリット・意図は省かれ、情報そのものと明確に切り離されている |

## 4 結果と考察

分析の結果，グループ因子数は A: 3 件，B: 3 件，C: 3 件，D: 2 件となった．A, B, C で因子数 4 以上の場合，$\omega_{total}$ が 0.6 以上と大きい因子が観測された．つまり，この因子の 60％以上が一般因子により二次的に説明できることとなり，独立性が低いと判断した．また A, C では $\omega_{group}$ の値が 0.1 未満となるような独立性の低いグループ因子も確認されたため，因子数 4 以上は不適と判断した．

評価指標に注目すると，$\omega_{total}$ は全メディアで 0.9 以上と，推定したモデルにより 9 割以上の観測変数の変動を説明できた．また ECV は D 以外では 0.4 前後と低い値となった．D のみ 0.6 以上と比較的高く，一般因子の消化しやすさへの影響が大きいことが分かった．グループ因子に対し広範な概念を指す一般因子が説明する割合が高くなったことで，適合度も大きく算出されたと考えられる．

具体的な解釈結果について，まずオンラインニュース記事（A）のグループ因子は親密性，未開拓性，簡潔性である．親密性は情報に対する事前知識の多さ，未開拓性は情報に対する事前知識・先入観の無さ，簡潔性は要点が集約度や要点以外の要素の少なさをそれぞれ意味する．また因子負荷が大きい観測変数群から，一般因子は未開拓性と他メディアでも抽出された網羅性を内包する概念だと解釈される．したがって，共通する情報量の多さを指す充実性と解釈した．また，親密性と未開拓性は定義や構成する観測変数群から正反対の因子だと分かる．ただし，これは矛盾ではなく，専門家であることと事前知識がないことは共に消化しやすさに寄与することを示していると考えられる．すなわち，中途半端な知識の所持が最も消化効率が悪い状態と見なされていると言える．なお，$\omega_{total}$ は親密性の方が高く，消化しやすさへの寄与は大きい．ただし，$\omega_{general}$ や因子負荷の値を参照するとの方が一般因子の影響が少ないため，一般因子の影響を除けば，親密である方が未開拓であるより望ましいと言える．

オンライン広告（B）のグループ因子は網羅性，簡潔性，結論へのアクセス性であった．いずれも抽出した因子の $\omega_{total}$ は概ね同程度だったが，最も高かったのは情報の項目やタイプ・補足情報の豊富さを

指す網羅性だった. 次に簡潔性, 結論へのアクセス性と続いた. 簡潔性の構成要素は概ね A の同名因子と構成要素が概ね同じとなった. 要点の理解しやすさを示す結論へのアクセス性は親密性に類似しているが, 情報の集約性を指す観測変数の因子負荷も高く, 他の因子として解釈した. また構成要素から, 親密性・簡潔性の共通項である消化コストの低さを一般因子の解釈として採用した. B では結論や要点・その説明が多方面かつ多様な形で, かつ簡潔に説明されていると最も消化しやすいと考えられる.

オンラインショッピング (C) のグループ因子は概ね A と一致したが, $\omega_{\text{total}}$ がほぼ均一という違いがあった. 一般因子は A と異なり網羅性・親密性を包括するため, より多くのニッチなニーズに応えられるという意味で充足性とした. なお, A 同様に親密性・未開拓性が抽出されたが, C では $\omega_{\text{total}}$ の差は軽微で, $\omega_{\text{group}}$ は未開拓性の方が高く, 一般因子に関わらず消化しやすさへの寄与が大きい. すなわち, C では多様な情報が含まれつつも個々は簡潔に整理されていることが望ましく, またトピックや説明には親密である方が良いが, 中途半端な知識よりは事前知識がない方が消化効率は良いと言える.

最後に論文/レポート (D) では, 説明可能性, 簡潔性が抽出された. 説明可能性は結論に至るまでの経緯や背景の妥当さ・理解しやすさを指す. また簡潔性は上記 3 メディアと共通で, 全メディアで抽出された唯一の因子となった. これら 2 因子は他メディアと比較して $\omega_{\text{total}}$ の値が大きく, 設定する因子数の少なさから関連の大きい観測変数も多かった. これは D の適合度の高さに寄与していると考えられる. なお一般因子は被読了性と解釈した. この因子との関連の強さが上位の観測変数は未開拓性の構成要素に類似しているが, 因子負荷が 0.60~0.70 と A や C よりも大きい. また 2 つのグループ因子の $\omega_{\text{general}}$ の値はそれぞれ 0.57, 0.65 と比較的高く, 一般因子と相関は高い. したがって, 未開拓性・説明可能性・簡潔性の全てが高い状態を包括する概念として解釈した. 以上から, D は結論や新規性が分かりやすく, その論旨展開や背景も含めて滞らずに一読できることが望ましいと分かった. これは論旨展開や背景等に重きが置かれるという D の特徴と直感的にも反しない. 他メディアよりも情報伝達の役割が明確で, 発信者・受信者間で前提として成立しているため, 適合度が大きくなったと考えられる.

## 5 まとめ

本研究では, 定量評価モデルと階層的因子分析の導入により, 共通の観測変数群よりメディア別に消化しやすさを構成する因子を抽出し, メディア内の共通要素やメディア間の共通因子の存在が示唆される結果を得た. メディアごとに評価項目は個々に異なるものの, 無秩序な情報量の増大は情報伝達本来の目的に対し共通して好ましくないことが分かった. また論文/レポートの適合度の高さに見られるように, メディアの伝達目的が発信者・受信者間で明確だと個々人の消化しやすさの指標のばらつきが収束する可能性が示唆された.

NLP2023 発表論文原稿の訂正について

言語処理学会第 29 回年次大会（NLP2023）発表原稿の本文について、下記のように訂正させていただきます。皆様に多大なるご迷惑をお掛けしてしまい、心よりお詫び申し上げます。

| 題目：テキスト・画像による情報消化効率のメディア別構成因子抽出 | | |
|---|---|---|
| 該当箇所 | （誤） | （正） |
| p.2, (2)式 | $\rho(A) = \text{ev}(親密性) \cdot 0.272 + \text{ev}(未開拓性) \cdot 0.360 + \text{ev}(簡潔性) \cdot 0.378$<br>$\rho(B) = \text{ev}(網羅性) \cdot 0.384 + \text{ev}(未開拓性) \cdot 0.360 + \text{ev}(簡潔性) \cdot 0.378$<br>$\rho(C) = \text{ev}(親密性) \cdot 0.272 + \text{ev}(簡潔性) \cdot 0.312 + \text{ev}(結論へのアクセス性) \cdot 0.304$<br>$\rho(D) = \text{ev}(説明可能性) \cdot 0.557 + \text{ev}(簡潔性) \cdot 0.443$ | $\rho(A) = \text{ev}(未開拓性) \cdot 0.272 + \text{ev}(簡潔性) \cdot 0.360 + \text{ev}(親密性) \cdot 0.378$<br>$\rho(B) = \text{ev}(網羅性) \cdot 0.384 + \text{ev}(簡潔性) \cdot 0.312 + \text{ev}(結論へのアクセス性) \cdot 0.304$<br>$\rho(C) = \text{ev}(親密性) \cdot 0.279 + \text{ev}(簡潔性) \cdot 0.400 + \text{ev}(未開拓性) \cdot 0.321$<br>$\rho(D) = \text{ev}(説明可能性) \cdot 0.557 + \text{ev}(簡潔性) \cdot 0.443$ |
| p.2, 3.3 節 | $\omega_{\text{total}}$ は一般因子が変動の何割を説明できたかを示す． | $\omega_{\text{total}}$ は全因子が変動の何割を説明できたかを示す． |
| p.3, 4 章 | なお，$\omega_t$ は親密性の方が高く，消化しやすさへの寄与は大きい．ただし，$\omega_{\text{general}}$ や因子負荷未開拓性の方が一般因子の影響が少ないため，一般因子の影響を除けば，未開拓である方が親密であるより望ましいと言える． | なお，$\omega_t$ は未開拓性の方が高く，消化しやすさへの寄与は大きい．ただし，$\omega_{\text{general}}$ や因子負荷や親密性の方が一般因子の影響が少ないため，一般因子の影響を除けば，親密である方が未開拓であるより望ましいと言える． |
| p.3, 4 章 | 以上を踏まえた因子の最終的な解釈結果が図 4 である．まずオンラインニュース記事（A）のグループ因子は親密性，未開拓性，簡潔性である． | 以上を踏まえた因子の最終的な解釈結果について，まずオンラインニュース記事（A）のグループ因子は親密性，未開拓性，簡潔性である． |
| p.4, 4 章 | この因子との関連の強さが上位の観測変数は未開拓性の構成要素に類似しているが，因子負荷が 0.60~0.70 と A や C よりも大きい（図 2）． | この因子との関連の強さが上位の観測変数は未開拓性の構成要素に類似しているが，因子負荷が 0.60~0.70 と A や C よりも大きい． |